# A low-cost laser scanning solution for flexible robotic cells: spray coating

M. Ferreira, A. P. Moreira

*Robotics and Intelligent Systems Group (ROBIS) - INESC Porto, University of Porto, Rua Dr. Roberto Frias, 378, 4200-465 Porto, Portugal*

Tel.: +351 222 094 398

Fax: +351 222 094 050

E-mail: marcos.ferreira@ieee.org

Url: http://gnomo.fe.up.pt/~robotic/

P. Neto

*Department of Mechanical Engineering (CEMUC) - POLO II, University of Coimbra, 3030-788 Coimbra, Portugal*

Tel.: +351 966 025 289

Fax: +351 239 790 701

E-mail: pedro.neto@dem.uc.pt

Url: http://robotics.dem.uc.pt/pedro.neto/

**Abstract:** In this paper, an adaptive and low-cost robotic coating platform for small production series is presented. This new platform presents a flexible architecture that enables fast/automatic system adaptive behaviour without human intervention. The concept is based on contactless technology, using artificial vision and laser scanning to identify and characterize different workpieces travelling on a conveyor. Using laser triangulation the workpieces are virtually reconstructed through a simplified cloud of 3D points. From those reconstructed models several algorithms are implemented to extract information about workpieces profile (pattern recognition), size, boundary and pose. Such information is then used to on-line adjust the "base" robot programs. These robot programs are off-line generated from a 3D computer-aided design (CAD) model of each different workpiece profile. Finally, the robotic manipulator executes the coating process after its "base" programs have been adjusted. This is a low-cost and fully autonomous system that allows adapting the robot's behaviour to different manufacturing situations. It means that the robot is ready to work over any piece at any time, and thus, small production series can be reduced to as much as a one-object-series. No skilled workers and large setup times are needed to operate it. Experimental results showed that this solution proved to be efficient and can be applied not only for spray coating purposes but also for many other industrial processes (automatic manipulation, pick-and-place, inspection, etc.).

# 1 Introduction

## 1.1 Motivation

Production lines tend to evolve into the concept of mass customization, i.e., working on small production series with flexible and customized procedures to each of them according to customers' demands. Consequently, this means that high flexibility and versatility are mandatory concepts to the production lines of today. However, the setup and reconfiguration time of those flexible manufacturing systems (many times robot-based systems) is still too large when compared with the effective production time. System reconfiguration and associated downtimes implies strong financial efforts. Moreover, highly qualified and skilled workers are needed to operate this kind of flexible systems. This is a problem since many companies have no budget to hire skilled workers.

Robotic manipulators are often a key element of flexible manufacturing systems. However, industrial manipulators still take a long time to be programmed. In fact, robot programming is a time consuming task that usually requires experienced and highly qualified workers to perform it. Despite these drawbacks, robotic manipulators are strongly desired in modern production lines. They have a series of advantages over human labour such as the ability to work continuously, high accuracy and repeatability, immunity to fatigue, immunity to distractions and the capacity to work in hazardous environments.

Taking the case study of an adaptive robotic coating system for small production series, this paper presents a low-cost and flexible architecture that enables fast system adaptation to changing conditions (product variants) without human intervention.

Most of the manufactured products need to be coated to improve their visual appearance and/or to provide protection from corrosion or damage. Manual coating operations can cause many problems such as environment pollution, coating material waste, inconsistent quality and low productivity [1]. These are some of the reasons why many companies are changing their manual coating systems to automated ones. This research work was initiated at the request of a small and medium-sized enterprise (SME) named FLUPOL. This SME is an industrial coatings applicator that usually works with small production series and with very different products (industrial bakeware, automotive parts, housewares, etc.).

**1.2 Proposed architecture and technologies**

The proposed platform integrates three different sub-platforms, Fig. 1. The artificial vision system captures images of the workpieces travelling along a conveyor, and on which a laser-line is projected (3D laser scanning). The laser-line scans the entire workpieces as they are transported (note that the camera and the laser are fixed while the workpiece is being transported on the conveyor). Then, using laser triangulation the workpieces are virtually reconstructed through a simplified cloud of 3D points. From those models several algorithms are implemented to extract information about workpieces size, boundary, profile and pose. A k-nearest neighbour (KNN) classifier is used to classify the different workpieces (pattern recognition).

Generally speaking, the proposed platform produces the 3D reconstruction of the workpieces, identifies those workpieces and also provides information about workpieces' pose. All this information enables on-line automatic system adaptation to the working scenario according to the specific profile and pose of each workpiece. In practice, this information is used to automatically select and adjust the "base" programs that run on the robot controller. These robot programs are off-line generated from a 3D CAD model of each different workpiece profile. Finally, after the selected "base" robot programs have been adjusted on-line, the robotic manipulator executes the coating process. Its coating schemes are adapted to match the workpieces' size and layout.

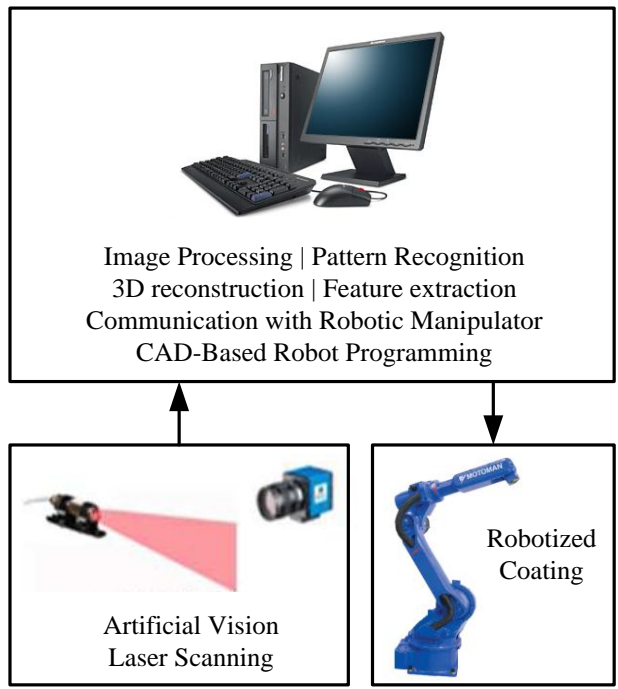

Fig. 1 Simplified platform architecture

## 1.3 Related work and discussion

The integration of artificial vision with laser triangulation, pattern recognition and flexible reprogramming schemes of industrial manipulators has not yet been intensively discussed in literature, at least for all these fields' together. On contrary, artificial vision, laser based scanning systems for different applications as well as pattern recognition techniques have been largely reported in literature.

Profile acquisition and recognition of real world objects is an important issue for modern manufacturing systems. Numerous technologies have been studied to perform the above mentioned tasks, all of them with a wide range of hardware costs and different levels of achievable accuracy and detail. Streaming video and image-based techniques, structured light and laser light-sectioning methods, time-of-flight range finders, shape-from-silhouette algorithms and space carving techniques are some of the methods which have been studied in recent years [2-3]. Moreover, machine learning techniques [4-6], modelling of error [7] and methods to deal with time delays [8-9] are often associated with this type of technologies.

In the last few years, several techniques related to laser scanning have been studied. This includes a scanning system to reconstruct a 3D surface as a large set of polygonal meshes [10], a scanning system for complex surfaces [11] and a CAD/scanner based framework for robotic coating of complex products where the surface models are generated from points cloud [1]. A stereo head of cameras for a triangulation-based laser sensor device has been used for object recognition purposes [12]. An image-based visual seam tracking system for butt weld of thin plates where a structured laser light is used to detect the welding torch deviation is proposed by Fang *et al.* [13]. Marshall *et al.* proposes a solution based on the segmentation of range images for 3D reconstruction purposes [14]. Lowe proposes a 3D object recognition system from single 2D images [15]. A laser-scan system for medical applications was proposed by Hayashibe *et al.* [16]. An approach to automatically generate robot programs for spray painting of unknown parts is presented by Vincze *et al.* [17]. This system is based on laser triangulation sensing, geometric feature detection, robot tool path planning and generation of collision-free robot programs. In fact, the goal of this system is similar to the goal of the system proposed in this paper. However, some substantial differences in methodology can be pointed out: the solution proposed in [17, 18] can be uneconomical for many companies since it uses some relatively expensive commercial solutions. Moreover, the process of 3D reconstruction is performed combining a large number of elementary geometries while in the system proposed in this paper the workpieces are classified and 3D reconstructed through a simplified cloud of points.

Owing to recent advances in laser scanning technology, the set of dense points collected from the surface of a physical object can contain millions of points (point cloud data), leading to significant computational challenges. In this way, point cloud simplification algorithms have been studied [1, 19]. The solution proposed in this paper addresses this situation by using simplification methods to reduce the number of necessary points to virtually reconstruct a certain object with a predefined level of accuracy. This makes the reconstruction process faster.

Looking at other fields of application of vision and laser based scanning, a lot of research has been carried out with facial recognition [20-21], dimensional measurement of objects [22] and even for inspection and control of quality purposes [23-24]. Kwok *et al.* proposes a laser based system to collect 3D data around the surface of a turbine blade. From the reconstructed blade model a tool path is generated [25]. Another similar system is dedicated for turbine blades repair through the reconstruction of the blades from multiple range images [26]. A recent study analyses the applicability of scanning systems for geometric and dimensional tolerance control [27]. In fact, some analysis on precision have already been made, for example comparing the use of single or multi laser beams [28], or, exploring alternative computer vision systems as stereoscopic pairs [29]. Aliakbarpour *et al.* presents a good coverage on calibrating camera-laser setups [30]. Another interesting approach reports an industrial application in which a robotic assembly of a car door is assisted by a laser scanning system [31].

As a final summary, we can point out that most of the existing systems (off-the-shelf and laboratory prototypes) similar to ours are complex to use, suffer from lack of portability and usually are highly expensive when compared to the custom setup presented in this paper.

## 2 Artificial vision and laser triangulation

The artificial vision subsystem is responsible for capturing images of the workpieces travelling along a conveyor. A fixed laser-line is projected onto those workpieces. This laser-line is identified in each video frame and thereby it will serve as input to generate three-dimensional information about each workpiece. Image processing begins with the isolation of the laser-line for each frame: the environment illumination is controlled (this makes the area the camera is filming dark) and this way the laser-line appears brighter in the images. A binarization algorithm (1) is applied to each image, allowing us to work over very clean images as the one in Fig. 2 ( $F_{B\&W}$ is the binarized image). The laser-line appears in white and everything else in black.

$$F_{B\&W}[u,v] = \begin{cases} 0 & if\ [u,v] < threshold \\ 1 & if\ [u,v] \geq threshold \end{cases} \quad (1)$$

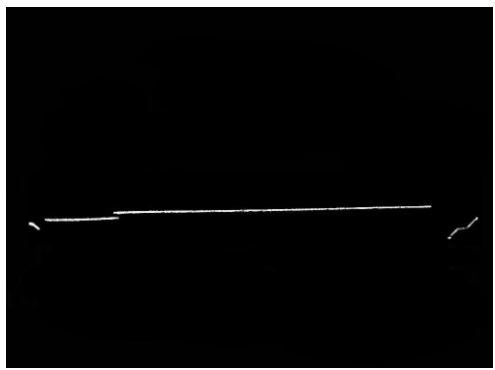

Fig. 2 Binarized image

## 2.1 Camera calibration

Camera calibration is one of the main issues of artificial vision systems. It allows identifying the artificial vision system parameters, in other words, calibration allows us to recognize the position in the world of any image pixel and vice-versa. Assuming that all coordinate systems are Cartesian, in homogeneous coordinates the relation between pixel coordinates $P_x$ and world coordinates $P$ can be seen in [32-33]. Thus, we have:

$$P_x = H \cdot P \qquad (2)$$

Where $H$ is the projection matrix whose fields are parameters we want to estimate, namely the translation and rotation of the camera, the focal length, the point where the optical axis cuts the image plane and the relation between sensor and pixel size. For this purpose, we have used OpenCV routines. These OpenCV routines are commonly used in the scientific community for camera calibration purposes. Along with the parameters mentioned above, these routines also provide information about the parameters for compensating the radial and tangential distortions of the images [34]. Calibration is achieved simply by showing to the camera different views of a chessboard pattern with known size.

## 2.2 Laser calibration

One might consider that the laser-line projected onto the workpiece actually originates a plane $L$, Fig. 3. The intersection of plane $L$ with the half-line obtained by $P = H^{-1} \cdot P_x$ results in a single and well defined point in space. Note that the laser must be placed obliquely to the camera. The equations of both the half-line $r$ and plane $L$:

$$r: P_r = P_{r0} + w_r \cdot t, \quad t \in \square \qquad (3)$$

$$L: w_n \cdot (P_L - P_{L0}) = 0 \qquad (4)$$

Where $P_{r0}$ and $P_{L0}$ are known points from the line and the plane respectively, for example the beginning of the half-line (position of the camera) and the position of the laser that also belongs to the plane; $w_r = [x_r \ y_r \ z_r]^T$ is the vector that contains the direction of the half-line and $w_n = \begin{bmatrix} x_n & y_n & z_n \end{bmatrix}^T$ is a vector orthogonal to the laser plane; $P_L$ is a point of the plane $L$. Finding the point that belongs simultaneously to $r$ and $L$, we have:

$$(P_{r0} + w_r \cdot t - P_{L0}) \cdot w_n = 0 \tag{5}$$

And then we get $t$ as:

$$t = \frac{d - x_n \cdot x_{r0} - y_n \cdot y_{r0} - z_n \cdot z_{r0}}{x_n \cdot x_r + y_n \cdot y_r + z_n \cdot z_r} \tag{6}$$

Where:

$$d = w_n \cdot P_{L0} \tag{7}$$

Supposing the camera had been calibrated before this step, the half-line equation is already known. Coming to know the laser plane parameters is also quite simple. We start by measure the laser position in the world and then we get two more points non-collinear with the laser position. Then, we just measure two points from the laser line when it hits any object in the world. After this, the laser is calibrated.

By replacing $t$ in (5) we get the world coordinates $P_r$ of any point of the image which also belongs to the laser-line. We can now jump, unequivocally, between image coordinates and world coordinates.

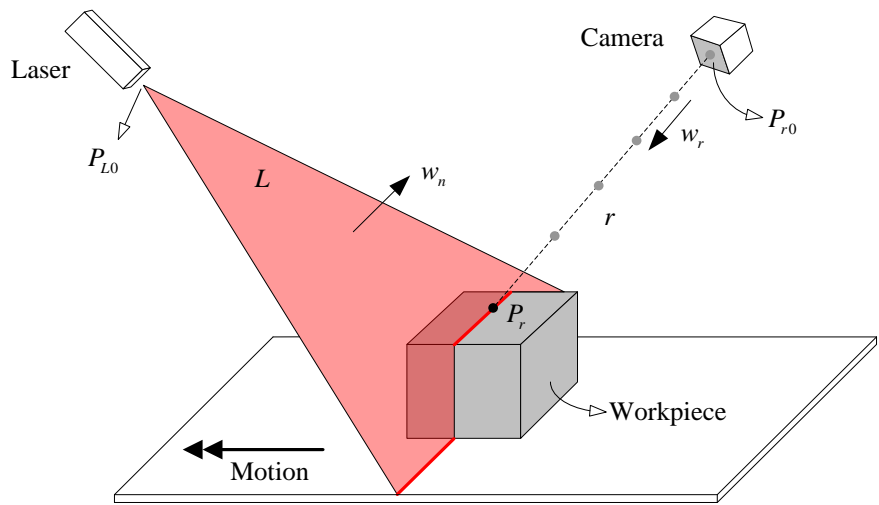

Fig. 3 Intersection of the laser plane with the half-line containing all the points that are projected into the same pixel

# 3 3D reconstruction

Virtual 3D reconstruction of real profiles has gained an increasing importance in industry. For the proposed approach, while the workpieces are being transported on the conveyor, the camera and laser setup keeps unaltered (fixed). Thus, owing to the workpieces motion, the form of the laser-line captured by the camera is changed according to the moving workpieces are gone. In fact, virtual representations of real objects give users a "feeling" about the real aspect of the reconstructed workpiece. In this way, these virtual models enable quick visual validation and errors may be tracked.

The real workpieces can be recreated in a 3D artificial environment by analysing the successive image frames and storing 3D data extracted from each one. Nevertheless, a more natural way to store collected 3D data is needed, other than having a set of loose points in space. Collected data should be well organized and treated by efficient algorithms. Hence, apart from visuals, all collected points were stored in a matrix form, making the indexes $i$ and $j$ of the matrix match with two of the axis of the world reference frame, Fig. 4. The third direction elements are associated with each pair of indexes ($i$ and $j$). In this way, one can map a certain volume in space into a simple matrix form. The points that describe the 3D models of a given workpiece are then arranged into a data structure that can be written as:

$$\begin{bmatrix} x \\ y \\ z \end{bmatrix}_w = \begin{bmatrix} 0 \\ y_{\min} \\ z_{\min} \end{bmatrix}_w + \begin{bmatrix} Matrix(i, j) \\ S_c \cdot i \\ -S_c \cdot j \end{bmatrix} \qquad (8)$$

Where $S_c$ is a scale factor that correlates the index length with its world length. $y_{\min}$ and $z_{\min}$ are just the offset between the matrix origin and the world origin. Note that one more time subscript $w$ indicates the world frame.

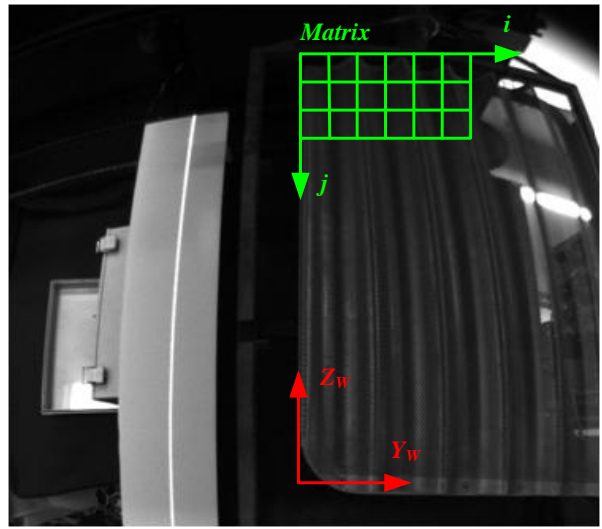

Fig. 4 Relationship between matrix indexes and world reference frame over a camera snapshot

## 3.1 Results

Fig. 6 presents two different views of a 3D reconstructed workpiece, the horizontal wavy-surface plate shown in Fig. 5, with approximately height = $0.80m$ and width = $0.40m$. These plates shown in Fig. 5 are examples of industrial bakeware produced by FLUPOL. Addressing the requirements of the proposed platform, the results of the reconstruction process are very good. Comparing the dimensions of the reconstructed workpieces with the real values, we can establish an error value. In fact, it is important to quantify the error we have on the acquisition of workpiece dimensions. Error is more prominent on the edges of the captured images (the edges of the workpieces) due to large barrel distortion. Consequently, shorter workpieces have a minor error. In the case of one of the biggest horizontal wavy-surface plates (80x40) we have an error of 2.2% on length and 2.1% on width. After several tests with different workpieces we can establish an average of 2% of error as an upper bound.

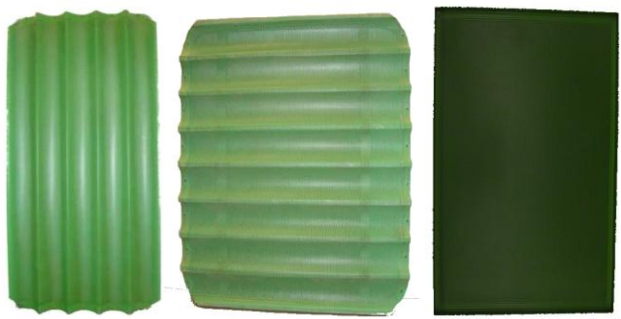

Fig. 5 Three different workpieces produced by FLUPOL. Horizontal wavy-surface plate (left), vertical wavy-surface plate (middle) and smooth surface (right)

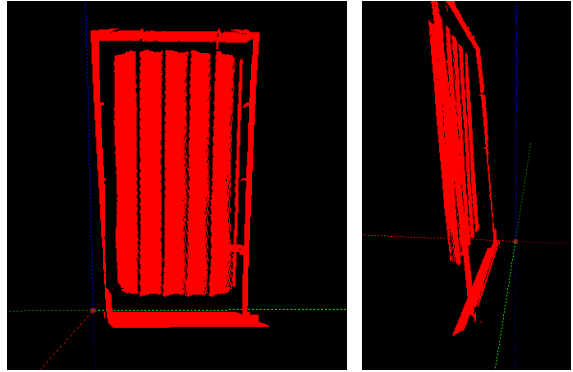

Fig. 6 3D reconstruction of the horizontal wavy-surface plate. Front view (left) and lateral view (right)

# 4 Feature extraction and pattern recognition

## 4.1 Border calculation

Using the matrix representation mentioned in previous section, a segmentation algorithm was implemented so that any workpiece profile could be evaluated. For the particular case of the coating process under study, all workpieces are transported on a rectangular metallic support (Fig. 13). This evidence contributed to facilitate the process of 3D reconstruction. Fig. 7 shows a flowchart reporting the implemented algorithm for workpiece border reconstruction purposes. Fig. 8 shows the result of applying the proposed border reconstruction method.

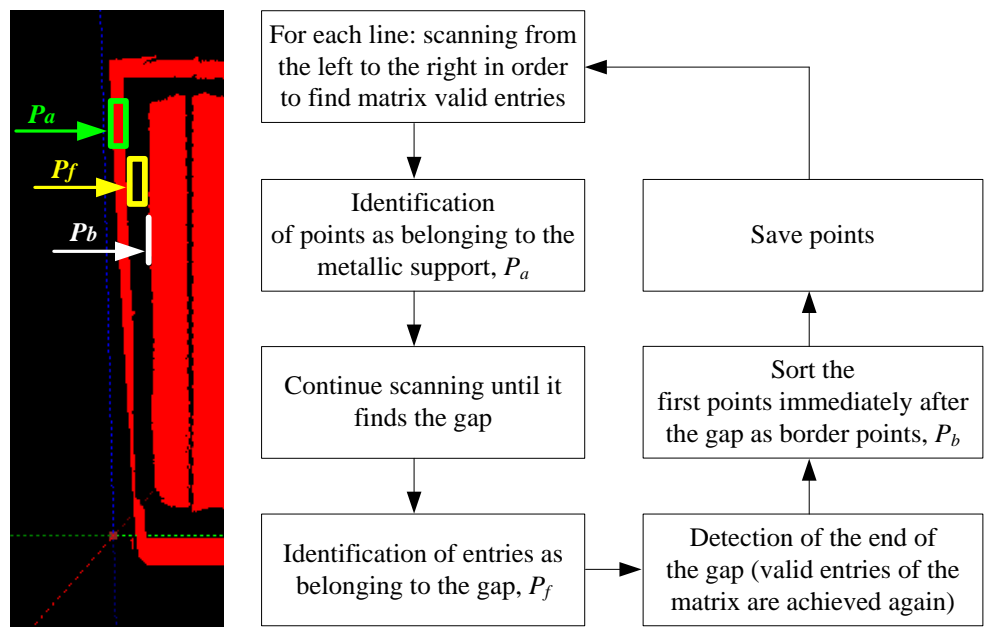

Fig. 7 Border reconstruction method

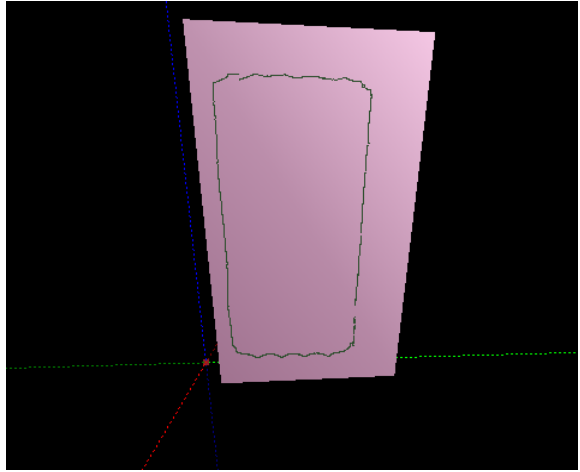

Fig. 8 Border points of a reconstructed workpiece

## 4.2 Slope calculation

As the metallic support that "supports" the workpiece is fixed to the conveyor in only two points, due to conveyor motion the hanging workpiece tends to deviate from its vertical pose. Thus, an important feature to extract from the reconstructed workpiece is the slope of the workpiece in relation to an "ideal" vertical pose. The proposed platform allows the estimation of workpiece inclination in a trivial way. In fact, the estimation of workpiece slope is an important issue when the coating phase is initiated. Other methods or techniques would most likely need additional sensors to the same end, making the whole setup more expensive and trickier to deal with.

Starting from a 3D reconstructed model and adequately choosing some model points, the plane that best fits those points can be computed. We are not using all of the model points because it would severely slow down the computations as we are talking about almost one million points. Thus, we are mostly using the edge points of the reconstructed models to define the plane. The first step is to eliminate outliers, points out of the border area and points with very different depth comparing to other points inside the border region. Then, the "clean" edge points are the input for the regression process. The plane is calculated by following a method that consists in solving the least squares problem using the singular value decomposition (SVD) technique [35].

Some factors affect the quality of the plane provided by the SVD analysis. There are problems both with outliers (above mentioned) and the non-symmetry of the reconstructed models. As we have only one laser (placed obliquely to the camera), when the wavy-like models are scanned some areas cannot be reached by the laser-line (hidden areas). As an example, with the camera orthogonal to the workpiece and the laser pointing from left to right, the right side of the waves cannot be scanned. The result is a non-symmetric 3D model. Then, this is another reason because we are mostly using the edge points of the reconstructed models as input for the SVD analysis. These are just some hundred points, no serious threat on computational time. For a sample of 1000 points, the SVD process takes about 10 milliseconds.

In terms of slope error, it is difficult to estimate it because it is substantially different for each different workpiece. The uncertainty comes from the way the operator places the workpieces in the conveyor.

Fig. 9 shows the plane that represents an approximation to a specific workpiece slope is superimposed on the reconstructed workpiece model. The orientation of this plane gives an approximation to the real workpiece deviation from the "ideal" vertical pose.

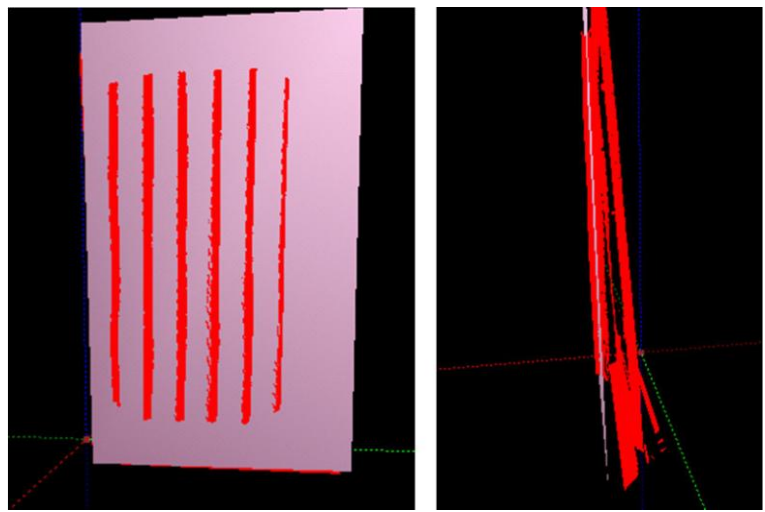

Fig. 9 Plane representing the workpiece slope, front view (left) and lateral view (right)

## 4.3 Pattern recognition

The proposed platform should be able to distinguish between at least three different types of workpieces: one with a smooth surface and two others with undulated surfaces (vertical or horizontal), Fig. 5. Each one of these types of workpieces can have varying dimensions (Table I). The proposed system should be able not only to recognize workpieces profile but also workpieces different sizes.

Table I: Horizontal wavy-surface plate model sizes

| **Model** | **Length (*cm*)** | **Width (*cm*)** |
|-----------|-------------------|------------------|
| Model #1  | 80                | 40               |
| Model #2  | 60                | 40               |
| Model #3  | 61.5              | 48               |
| Model #4  | 75.5              | 48               |
| Model #5  | 56                | 38               |
| Model #6  | 55                | 36.5             |

The recognition/classification of the workpieces profile is carried out using a KNN classifier [4]. An *N*-dimensional space is created extracting *N* features from the 3D reconstructed models. Some workpieces are used for training purposes, in other words, the *N* features are off-line computed and then manually classified. When the system is running, upon detecting a new workpiece, the workpiece model is created, features are extracted and then the distance to each one of the training points is computed (within the feature space). Choosing the k-nearest points, the workpiece under analysis will be attributed to the most common class among the k-neighbors.

For this type of industrial application we cannot expect less than an almost "perfect" classifier, i.e., one that would return near 100% of accurate classification. Since after the classification process a robot program is uploaded to the robot controller, a wrong classification could lead to dangerous situations (collisions) and/or wasting of coating product. To achieve such a classifier, instead of computing a lot of common features and proceed to more complex classifiers, research focused on the search for "good" features. These "good" features will allow to distinguish the different workpieces in a direct and easy way. The selected features were $\delta^2_{mHoriz}$ and $\delta^2_{mVert}$, both representing the variance of depth in *M* slices of the 3D models (in horizontal and vertical direction respectively). The 3D models are sliced horizontally and vertically as Fig. 10 suggests. Therefore, each $\delta^2_m$ stands for the mean of the variance on depth in those cuts, which themselves are built up from $N_c$ points. In this way, $\delta^2_m$ is defined as:

$$\delta_m^2 = \frac{1}{M} \sum_{j=1}^{M} \left( \frac{1}{N_c} \sum_{i=1}^{N_c} \left( x_i^j - x_P^j \right)^2 \right) \quad (9)$$

Where $x_i^j$ is a point in cut $j$ and $x_P^j$ is the mean of points in cut $j$.

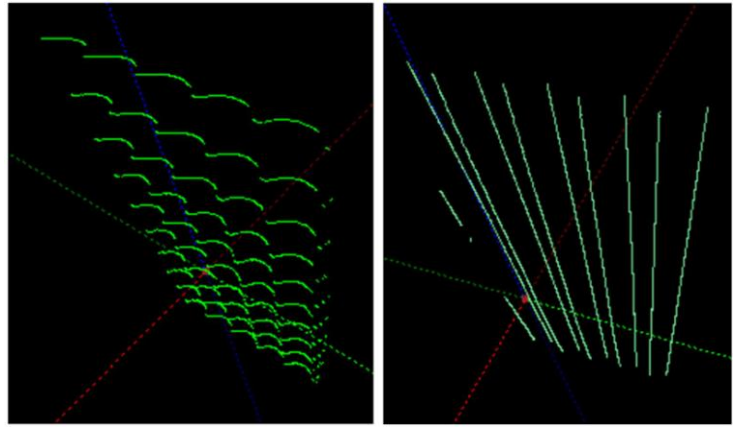

Fig. 10 Sliced representations of the horizontal wavy-surface plate; horizontal cuts (left) and vertical cuts (right)

### 4.3.1 Results and discussion

Results showing the reason for having near 100% of accurate classification are presented in Table II and summarized in Fig. 11. It can easily be seen that for smooth surfaces, both $\delta_{mHoriz}^2$ and $\delta_{mVert}^2$ take small values. On the other hand, an undulated surface presents high depth variance according to the direction of the "waves", i.e., an horizontal wavy profile has high $\delta_{mHoriz}^2$ and low $\delta_{mVert}^2$, and vice-versa when thinking about a vertical wavy profile. Results presented in Table II are average values calculated from five tests for each different type of workpiece profile.

Table II: Features values

| **Features** | $\delta_{mHoriz}^2$ | $\delta_{mVert}^2$ |
|---|---|---|
| Smooth surface | 0.7 | 0.8 |
| Horizontal wavy-surface | 5.8 | 1.4 |
| Vertical wavy-surface | 1.1 | 6.3 |

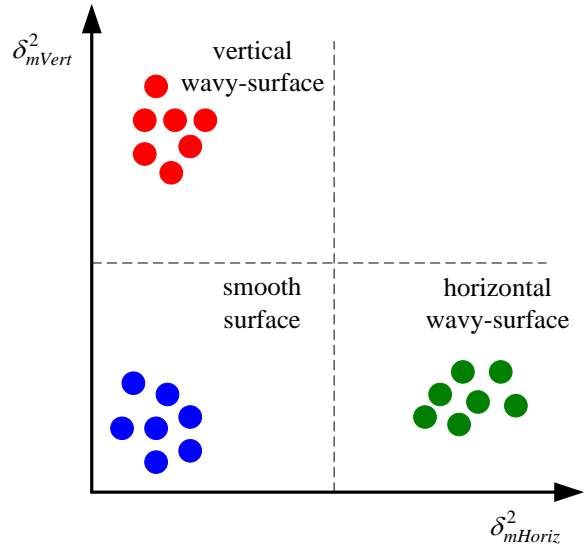

Fig. 11 Workpieces features

With respect to a more general application where there is no obvious distinction of objects with such a few features, more complex and robust algorithms other than KNNs may be applied. Classification algorithms based on artificial neural networks (ANNs) [5] or support vector machines (SVMs) are plausible solutions. However, these solutions would require a greater number of training samples. As extensively shown in the literature [36-38], designing this kind of classifiers with very few training samples brings additional problems. The design of the classifier is much more complex and high performance rates may not be achievable.

The implemented KNN classifier provides very good results. The implementation is simple and, if needed, we can always increase the training set later. This can be done without changing code and having to run the training again, as in the case of ANNs or SVMs. Adding training samples in KNN means just more computation when testing. Even for the application proposed in this paper, it may be necessary to classify other type of workpieces. The proposed solution is flexible and expansible enough to deal with that situation, simply reinventing and exploring new and "good" features to extract from the models.

Another totally different technique that could be implemented eying the same goal of recognizing objects is, for example, the attachment of RFID tags in each workpiece. This technique allows classifying the workpieces, but it requires that labels are manually placed in each different workpiece (different in profile and size). Moreover, with RFID tags we could not estimate the workpiece deviation from its "ideal" pose. The proposed platform allows us to compensate and deal with such negative issues: Otherwise, we will be forced to introduce new hardware in the system such as transporting guides to maintain workpieces' position and orientation constant.

# 5 Robot programming from CAD

The "base" robot programs are off-line generated from a 3D CAD model of each different workpiece (different in profile and not in size). Robot programs are designed so that robot motion is parameterized with the workpieces dimensions. These programs are kept in the robots' controller and called when needed.

Once CAD technology is today common throughout industry, any user with basic CAD skills can be able to generate robot programs off-line from a CAD model. In addition, the 3D CAD package (Autodesk Inventor) that interfaces with the user is a well known CAD package, widespread in the market at a relative low-cost. This system works as a real human-robot interface where, through the CAD, the user generates programs for the real robot. The methods used to extract information from the CAD models and techniques to treat/convert it into robot commands have been investigated and successfully implemented [39]. The information needed to program the robot (generation of coating gun trajectories) is extracted from the 3D CAD model of each different workpiece and from the virtual robot paths that the user can easily define in the CAD drawing, Fig. 12.

In order to achieve uniform coat thickness, the spatial gun position, orientation and velocity should be planned based on the local geometry of the free-form surface [40]. In this case study, it is desired that the plates can be coated according to their surfaces, e.g., horizontal-wavy profiles should be coated with vertical moves. Experiments were conducted to evaluate the interface performance, and results showed that the CAD-based system is easy to use and within minutes an untrained user can generate a robot program for a new workpiece.

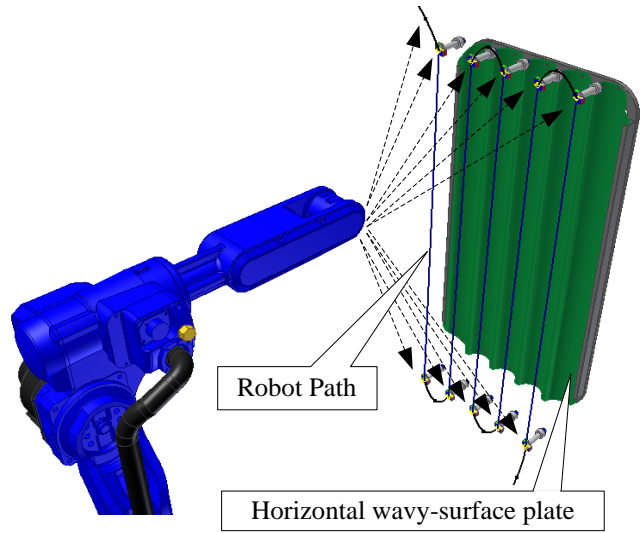

Fig. 12 A robot program will be generated from this CAD model

## 6 Experimental setup, results and future work

Results and critical analysis of the 3D reconstruction process, pattern recognition and feature extraction have been presented and discussed in previous sections. This section aims to present an overview of the entire system setup and its working performance. Fig. 13 shows the system setup where you can see a workpiece being transported, the laser-line projected onto the workpiece, the camera capturing images and a MOTOMAN HP6 robotic arm equipped with an NX100 controller. Note that both the laser (wavelength = 650*nm*, power = 3*mW*) and camera (Imaging Source DMK 31BU03 [41]) are fixed. The distance between the laser and the camera is 50*cm*. Fig. 14 presents a detailed architecture of the platform, where the off-line and on-line processes are highlighted. The ambient light is controlled to be dark. In this way, the laser-line appears brighter in the images.

Once a new workpiece has been identified by the system, multiple commands are immediately sent to the robot controller. These commands contain information about the type of workpiece (a "base" robot program is selected), workpiece size and the necessary slope adjustments. It means that, after the scanning process the robot receives the necessary information to perform the coating process in an appropriate way. A video is a good way to visualize the entire system setup and *modus operandi* [42]. The main quantitative results of the entire system are listed below:

- In the process of 3D reconstruction we have an average of 2% of error as an upper bound.
- In terms of classification we have near 100% of accurate classification.
- The scanning time is imposed by the velocity of the conveyor. For this particular process the conveyor velocity is 1*m/min*.

These results are in line with the results obtained by other similar studies (see section 1.3). However, the proposed platform has some important features that should be highlighted:

- The workpiece profile is reconstructed using a simplified cloud of points. Other studies use different methods for the same purpose [1, 17, 18].
- The proposed platform not only classifies workpieces profiles but also provides information on the size and slope of the workpiece under analysis. This is very important for the adaptive behaviour that characterizes the platform.
- The "base" robot programs are off-line generated from a 3D CAD model running on a commercial 3D CAD package. This is an alternative to commercial computer-aided robotics (CAR) software, which can be an expensive solution for many companies. This point reinforces the low-cost character of the proposed platform.

The greatest difficulties we have encountered throughout the research were:

- The calculation of workpieces slope.
- The development of a system able to deal with any workpiece profile and size.
- The achievement of "good" features for the classification phase. These features should be simple and size-invariant.

Generally speaking, the proposed platform works well and meets the requirements previously defined. We can say that the platform is flexible enough to adapt itself to each workpiece conditions. The classification algorithm needs further validation and development in order to make it more robust and comprehensive. This is important, especially in the definition of the features to extract from the workpieces, other than those presented here.

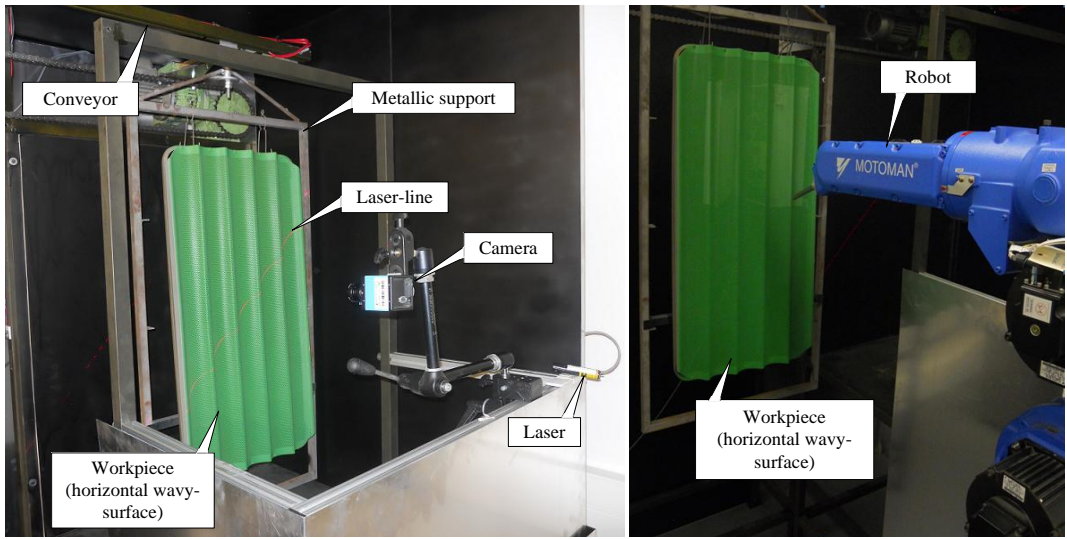

Fig. 13 System setup; laser scanning (left) and robotic coating process (right)

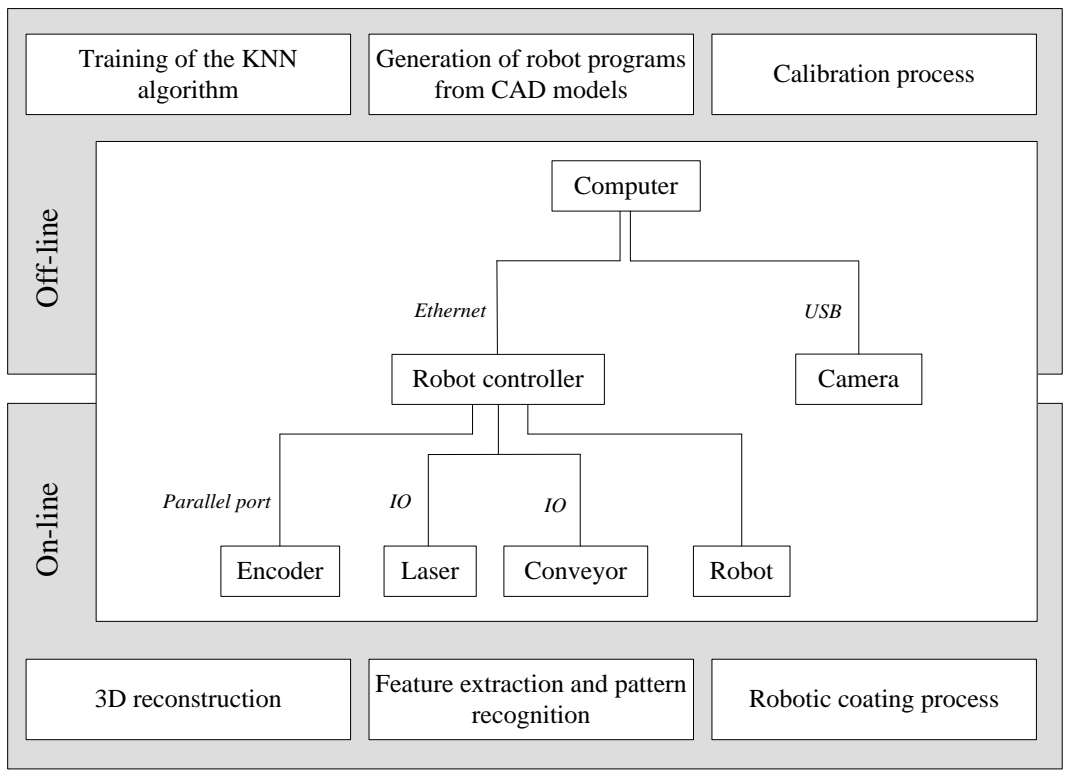

Fig. 14 System architecture. Off-line and on-line processes are highlighted

## 7 Conclusions

An adaptive scheme for an industrial coating process of small series was studied and developed. This is a low-cost and fully autonomous system that allows adapting the robot's behaviour to different manufacturing situations. It means that the robot is ready to work over any piece at any time, and thus, small production series can be reduced to as much as a one-object-series. No skilled workers and large setup times are needed to operate it.

The proposed solution is a noncontact scanning system where artificial vision together with laser triangulation allows an accurate 3D reconstruction of workpiece models. From those reconstructed models features are extracted and workpieces are classified with near 100% of accurate recognition. Base robot programs are generated off-line from 3D CAD models of the workpieces. These programs are automatically adjusted with information from the reconstructed workpiece models, adapting the robot coating movements to the workpieces size and pose.

Experimental results showed that the proposed solution proved to be efficient. Moreover, it can be applied not only for spray coating purposes but also for many other industrial processes where workpieces need to be recognized before robot(s) working on them. For example, this solution can be used for many different automatic manipulation processes or for inspection purposes. Most of the existing systems similar to ours (off-the-shelf and laboratory prototypes) are complex to use, suffer from lack of portability and usually are highly expensive when compared to the custom setup presented in this paper.

## References

[1] Bi ZM, Lang SYT (2007) A framework for CAD and sensor-based robotic coating automation. IEEE Trans Ind Inform 3(1): 84-91. doi: 10.1109/TII.2007.891309

[2] Bernardini F, Rushmeier H (2002) The 3D model acquisition pipeline. Comput Graphics Forum 21(2): 149-172. doi: 10.1111/1467-8659.00574

[3] Blais F (2004) Review of 20 years of range sensor development. J Electron Imaging 13(1): 231-240. doi: 10.1117/1.1631921

[4] Cover T, Hart P (1967) Nearest neighbor pattern classification. IEEE Trans Inf Theory 13(1): 21-27. doi: 10.1109/TIT.1967.1053964

[5] Neto P, Pires JN, Moreira AP (2010) High-level programming and control for industrial robotics: using a hand-held accelerometer-based input device for gesture and posture recognition. Ind Robot: Int J 37(2): 137-147. doi: 10.1108/01439911011018911

[6] Neto P, Pires JN, Moreira AP (2009) Accelerometer-Based Control of an Industrial Robotic Arm. Proc 18th IEEE Int Symp Robot Hum Interact Commun, pp 1192-1197. doi: 10.1109/ROMAN.2009.5326285

[7] Chen CW (2009) Modeling and control for non-linear structural systems via a NN-based approach. Expert Syst Appl: Int J 36(3): 4765-4772. doi: 10.1016/j.eswa.2008.06.062

[8] Hsiao FH, Chen CW, Liang YW, Xu SD, Chiang WL (2005) T-S fuzzy controllers for non-linear interconnected systems with multiple time delays. IEEE Trans Circuit Syst I 52(9): 1883-1893. doi: 10.1109/TCSI.2005.852492

[9] Cao YY, Frank PM (2000) Analysis and synthesis of non-linear time-delay systems via fuzzy control approach. IEEE Trans Fuzzy Syst 8(2): 200-211. doi: 10.1109/91.842153

[10] Borghese NA, Ferrigno G, Baroni G, Pedotti A, Ferrari S, Savarè R (2002) Autoscan: A flexible and portable 3D scanner. IEEE Comput Graphics Appl 18(3): 38-41. doi: 10.1109/38.674970

[11] Seokbae Son, Seungman Kim, Lee KH (2003) Path planning of multi-patched freeform surfaces for laser scanning. Int J Adv Manuf Technol 22: 424-435. doi: 10.1007/s00170-002-1502-0

[12] Aristos D, Tzafestas S (2009) Simultaneous Object Recognition and Position Tracking for Robotic Applications. Proc 2009 IEEE Int Conf Mechatron, pp 1-7. doi: 10.1109/ICMECH.2009.4957198

[13] Fang Z, Xu D, Tan M (2010) Visual seam tracking system for butt weld of thin plate. Int J Adv Manuf Technol 49(5-8): 519-526. doi: 10.1007/s00170-009-2421-0

[14] Marshall D, Lukacs G, Martin R (2001) Robust segmentation of primitives from range data in the presence of geometric degeneracy. IEEE Trans Pattern Anal Mach Intell 23(3): 304-314. doi: 10.1109/34.910883

[15] Lowe DG (1987) Three-dimensional object recognition from single two-dimensional images. Artif Intell 31(3): 355-395. doi: 10.1016/0004-3702(87)90070-1

[16] Hayashibe M, Suzuki N, Nakamura Y (2006) Laser-scan endoscope system for intraoperative geometry acquisition and surgical robot safety management. Méd Image Anal 10(4): 509-519. doi: 10.1016/j.media.2006.03.001

[17] Vincze M, Pichler A, Biegelbauer G, Hausler K, Anderson H, Madsen O, Kristiansen M (2002) Automatic robotic sparay painting of low volume high variant parts. Proc 33rd Int Symp Robot.

[18] Pichler A, Vincze M, Anderson H, Madsen O, Hausler K (2002) A method for automatic spray painting of unknown parts. Proc IEEE Int Conf Robot Automat, pp 444-449. doi: 10.1109/ROBOT.2002.1013400

[19] Hao Song, Hsi-Yung Feng (2009) A progressive point cloud simplification algorithm with preserved sharp edge data. Int J Adv Manuf Technol 45: 583-592. doi: 10.1007/s00170-009-1980-4

[20] Acosta D, García O, Aponte J (2006) Laser triangulation for shape acquisition in a 3D scanner plus scan. Proc 2006 Electron, Robot Automot Mech Conf, pp 14-19. doi: 10.1109/CERMA.2006.54

[21] Strat AV, Oliveira MM (2003) A point-and-shoot color 3D camera. Proc 4th Int Conf 3-D Digit Imaging Model, pp 483-490. doi: 10.1109/IM.2003.1240285

[22] Demeyere M, Rurimunzu D, Eugne C (2004) Diameter measurement of spherical objects by laser triangulation in an ambulatory context. IEEE Trans Instrum Measurement 56(3): 867-872. doi: 10.1109/TIM.2007.894884

[23] Noll R, Krauhausen M (2003) Multi-beam laser triangulation for the measurement of geometric features of moving objects in production lines. Proc 2003 Conf Lasers Electro-Optics Europ, pp 468. doi: 10.1109/CLEOE.2003.1313531

[24] Vezzetti E (2009) Computer aided inspection: design of customer-oriented benchmark for noncontact 3D scanner evaluation. Int J Adv Manuf Technol 41: 1140-1151. doi: 10.1007/s00170-008-1562-x

[25] Kwok KS, Louks CS, Driessen BJ (1998) Rapid 3-D digitizing and tool path generation for complex shapes. Proc IEEE Int Conf Robot Automat, pp 2789-2794. doi: 10.1109/ROBOT.1998.680468

[26] Sheng X, Krömker M (1998) Surface reconstruction and extrapolation from multiple range images for automatic turbine blades repair. Proc IEEE IECON 1998 Conf, pp 1315-1320. doi: 10.1109/IECON.1998.722840

[27] Martínez S, Cuesta E, Barreiro J, Álvarez B (2010) Analysis of laser scanning and strategies for dimensional and geometrical control. Int J Adv Manuf Technol 46: 621-629. doi: 10.1007/s00170-009-2106-8

[28] Lei W, Mei B, Jun G, ChunSheng O (2006) A novel double triangulation 3D camera design. Proc 2006 IEEE Int Conf Inf Acquis, pp 877-882. doi: 10.1109/ICIA.2006.305849

[29] Ma X, Hongbin Z (2007) Hybrid scene reconstruction by integrating scan data and stereo image pairs. Proc 6th Int Conf 3-D Digit Imaging Mode, pp 393-400. doi: 10.1109/3DIM.2007.28

[30] Aliakbarpour H, Núñez P, Prado J, Khoshhal K, Dias J (2009) An efficient algorithm for extrinsic calibration between a 3D laser range finder and a stereo camera for surveillance. Proc 14th Int Conf Adv Robot

[31] Zussman E, Schuler H, Seliger G (1994) Analysis of the geometrical features detectability constraints for laser-scanner sensor planning. Int J Adv Manuf Technol 9: 56-64

[32] Nixon M, Aguado A (2008) Feature extraction and image processing. Elsevier, Oxford

[33] Savii GG (2004) Camera calibration using compound genetic simplex algorithm. J Optoelectron Adv Mater 6(4): 1255-1261

[34] Weng J, Cohen P, Herniou M (1992) Camera calibration with distortion models and accuracy evaluation. IEEE Trans Pattern Anal Mach Intell 14(10): 965-980. doi: 10.1109/34.159901

[35] Eberly D (2008) Least squares fitting of data. Geometric Tools LLC

[36] Mehrotra KG, Mohan CK, Ranka S (1991) Bounds on the number of samples needed for neural learning. IEEE Trans Neural Netw 2(6): 548-558. doi: 10.1109/72.97932

[37] Uchimura S, Hamamoto Y, Tomita S (1995) Effects of the sample size in artificial neural network classifier design. Proc IEEE Int Conf Neural Netw, pp 2126-2129. doi: 10.1109/ICNN.1995.489006

[38] Hamamoto Y, Uchimura S, Kanaoka T, Tomita S (1993) Evaluation of artificial neural network classifiers in small sample size situations. Proc 1993 Int Joint Conf Neural Netw, pp 1731-1735. doi: 10.1109/IJCNN.1993.716988

[39] Neto P, Pires JN, Moreira AP (2010) CAD-based off-line robot programming. Proc 4th IEEE Int Conf Robot, Autom Mechatron, pp 516-521. doi: 10.1109/RAMECH.2010.5513141

[40] Hertling P, Hog L, Larsen L, Perram JW, Petersen HG (1996) Task Curve Planning for Painting Robots - Part I: Process Modeling and Calibration. IEEE Trans Robot Autom 12(2): 324-330. doi: 10.1109/70.488951

[41] ImagingSource (2011) DMK 31BU03 camera datasheet. http://www.theimagingsource.com/en_US/products/cameras/usb-ccd-mono/dmk31bu03. Accessed 1 June 2011

[42] Video (2011) INESC Porto (University of Porto) and University of Coimbra. http://robotics.dem.uc.pt/pedro.neto/GS6.wmv. Accessed 1 June 2011